\documentclass[10pt,journal,compsoc]{IEEEtran}

\usepackage{enumitem}
\usepackage{booktabs}
\usepackage{mathtools}
\usepackage{multirow}
\usepackage{tabulary}
\usepackage{csquotes}
\usepackage{booktabs}
\usepackage{makecell}
\usepackage{color}

\ifCLASSOPTIONcompsoc
  \usepackage[nocompress]{cite}
\else
  \usepackage{cite}
\fi
\ifCLASSINFOpdf
\else
\fi
\begin{document}
%
\title{Looking At The Body: Automatic Analysis of Body Gestures and Self-Adaptors in Psychological Distress}
%
%
%
%

\author{Weizhe Lin,~\IEEEmembership{Member,~IEEE,}
        Indigo Orton,
        Qingbiao Li,
        Gabriela Pavarini,
        and Marwa Mahmoud,~\IEEEmembership{Member,~IEEE}
\IEEEcompsocitemizethanks{\IEEEcompsocthanksitem W. Lin is with the Department of Engineering, Cambridge.\protect\\
E-mail: wl356@cam.ac.uk
\IEEEcompsocthanksitem I. Orton, Q. Li and M. Mahmoud are with the Department of Computer Science and Technology, Cambridge.\protect\\Email: \{indigo.orton@cl., ql295@, mmam3@\}cam.ac.uk
\IEEEcompsocthanksitem G. Pavarini is with the Department of Psychiatry, Oxford.\protect\\Email: gabriela.pavarini@psych.ox.ac.uk
}
}

%
%

\markboth{Journal of \LaTeX\ Class Files,~Vol.~14, No.~8, August~2015}%
{Shell \MakeLowercase{\textit{et al.}}: Bare Advanced Demo of IEEEtran.cls for IEEE Biometrics Council Journals}
%



\IEEEtitleabstractindextext{%
\begin{abstract}
Psychological distress is a significant and growing issue in society. 
Automatic detection, assessment, and analysis of such distress is an active area of research. 
Compared to modalities such as face, head, and vocal, research investigating the use of the body modality for these tasks is relatively sparse. 
This is, in part, due to the limited available datasets and difficulty in automatically extracting useful body features. 
Recent advances in pose estimation and deep learning have enabled new approaches to this modality and domain. 
To enable this research, we have collected and analyzed a new dataset containing full body videos for short interviews and self-reported distress labels.
We propose a novel method to automatically detect self-adaptors and fidgeting, a subset of self-adaptors that has been shown to be correlated with psychological distress. 
We perform analysis on statistical body gestures and fidgeting features to explore how distress levels affect participants' behaviors.
We then propose a multi-modal approach that combines different feature representations using Multi-modal Deep Denoising Auto-Encoders and Improved Fisher Vector Encoding.
We demonstrate that our proposed model, combining audio-visual features with automatically detected fidgeting behavioral cues, can successfully predict distress levels in a dataset labeled with self-reported anxiety and depression levels. 
\end{abstract}

}

\maketitle

\IEEEdisplaynontitleabstractindextext

%
\IEEEpeerreviewmaketitle

\ifCLASSOPTIONcompsoc
\IEEEraisesectionheading{\section{Introduction}\label{sec:introduction}}
\else
\section{Introduction}
\label{sec:introduction}
\fi
\IEEEPARstart{P}sychological distress and mental disorders are significant threats to global health \cite{vos2017global}.\footnote{This work is an extension of the work in \cite{lin2020auto}, originally published in the proceedings of the IEEE International Conference on Automatic Face and Gesture Recognition (FG) 2020}
According to the World Health Organization (WHO), an estimated 450 million people around the world suffer from neuropsychiatric conditions \cite{world2001world}, with depression and anxiety being the most common mental disorders \cite{world2017world}.
Despite existing strategies for the treatment of distress, such as depression, it is estimated that nearly two-thirds of people suffering distress have never received help from a health professional \cite{mentalhealthuk}.
Early detection of distress is consistently noted as a key factor in treatment and positive outcomes.
Early detection requires an ongoing assessment to identify distress when it begins.
Self-evidently, ongoing assessment at scale is prohibitive when performed manually.
As such, automatic detection of signs of psychological distress or specific mental disorders is an active area of research.

Currently, the most effective automated distress detection approaches utilize multi-modal machine learning.
These modalities include facial, head, eye, linguistic (textual), vocal, and body. 

There are significant challenges to body modality research, particularly within automatic distress detection, including the lack of relevant data, the inability to share much of the data, and the difficulty in gathering such data.
Specifically, the combination of full-body data (either sensor-based or video-based) with psychological distress labels is rare.
Compounding this rarity is the private and sensitive nature of the data, which means such datasets are rarely shared publicly.

Body expressions, and especially self-adaptors, have been shown to be correlated with human affect, depression and psychological distress \cite{de2009bodies, neff2011don, mahmoud2016towards, chui2018semantic, chan2016behave}.
Self-adaptors are self-comforting gestures, including any kind of touching on other parts of the body, either dynamically or statically \cite{mahmoud2013automatic, fairbanks1982nonverbal}. Fidgeting, a subset of self-adaptors, is the act of moving about restlessly, playing with one's fingers, hair, or personal objects in a way that is not peripheral or nonessential to ongoing tasks or events \cite{mehrabian1986analysis}.
Patients with depression often engage in self-adaptors \cite{ekman1969repertoire}.
Fidgeting has been seen and reported in both anxiety and depression \cite{fairbanks1982nonverbal}; It is a sign of attention-deficit and hyperactivity disorder, also exhibited by individuals with autism \cite{froiland2016home}.
With manually annotated data, Scherer \textit{et al.} \cite{scherer2013automatic} reported a longer average duration of self-adaptors as well as fidgeting for distressed participants.

More recent advances in the state-of-the-art for pose estimation~\cite{cao2018openpose} enable accurate pose data on a broader set of datasets and thus open the door for new approaches for body expression analysis and broader incorporation of body features in multi-modal systems. 

In this paper, we propose to use a hierarchical model to automatically detect self-adaptors as well as fidgeting, which has been shown to be predictive of psychological distress. We analyzed body gestures and self adaptors in a dataset of video recordings that we collected, concentrating on symptoms of depression and anxiety because these are the most common mental disorders \cite{world2017world}.
We then present two methods to explore the body modality (especially fidgeting): First with a statistical linearity analysis with traditional linear regression, and second with a deep-learning-based pipeline.
In the second method, a Multi-modal Deep Denoising Auto-Encoder (multi-DDAE) is utilized for encoding per-frame features. Improved Fisher Vector encoding \cite{perronnin2010improving} is then used to generate per-sample representation. Finally, we demonstrate that these features are discriminative in psychological distress detection.

The contributions of this paper can be summarized as follows:

1) We introduce a new audio-visual dataset containing recordings of non-clinical interviews along with distress labels from established psychological evaluation questionnaires.

2) We propose a hierarchical model for automatic detection of self-adaptors (including fidgeting) from visual data and evaluate our approach on a publicly available fidgeting dataset with manual labels.

3) We present a statistical analysis of a set of statistical body gesture features as well as specific fidgeting features extracted from the body modality data and explore how distress levels affect participants' behavior in our dataset.

4) As proof of concept, we implement a multi-modal feature fusion framework to perform distress classification and demonstrate the importance of self-adaptors features, specifically fidgeting, in predicting symptoms of depression and anxiety.



\section{Related Work}\label{sec:related_work}
In this section, we focus on related work on automatic detection of signs of psychological distress, including studies that focus on separate modalities and multi-modal fusion frameworks.

\subsection{Facial and head modality}
Facial Action Coding System (FACS) \cite{friesen1978facial} has long been used to taxonomize human facial movements by their appearance on the face, which yields the concept of Facial Action Units (AUs). For example, the Audio/Visual Emotion Challenge (thereafter AVEC) used AUs features as a basic descriptor for its psychological distress detection tasks.

A big body of literature has been developed to analyze facial expressions and the head modalities in the context of depression and psychological distress. For example, Yang \textit{et al.} \cite{yang2017multimodal} proposed a ``Histogram of Displacement Range (HDR)'', which is a measurement of the amount of facial landmark movements to predict depression. Joshi \textit{et al.} \cite{joshi2013can} presented a categorization analysis framework which consists of ``bag of facial dynamics'' and ``histogram of head movements''. Dibeklio{\u{g}}lu \textit{et al.} \cite{dibekliouglu2015multimodal} \cite{dibekliouglu2017dynamic} feature-engineered dynamic representation (e.g., velocity, acceleration, and standard deviation of motion) for facial landmark movement and head motion and used them in a multi-modal system to detect depression in a dataset of clinical interviews.

Psychomotor retardation refers to a slowing-down of thought and a reduction of physical movements in an individual. Sobin \textit{et al.} \cite{sobin1997psychomotor} demonstrated the correlation between psychomotor retardation and depression. Syed \textit{et al.} \cite{syed2017depression} handcrafted descriptors using craniofacial movements in order to capture the psychomotor retardation, and then made predictions of depression.

Some other features such as
lower emotional expressivity \cite{scherer2014automatic}, 
eye lid movement \cite{syed2017depression}, reduced gaze activity \cite{alghowinem2015cross} \cite{anis2018detecting}, and averted gaze \cite{scherer2014automatic} have been also used as predictive features of depression.




\subsection{Audio modality}
Acoustic features of speech can be predictive of distress irrespective of the speech content \cite{ozdas2000analysis, srimadhur2020end}. For example, Ozdas \textit{et al.} \cite{ozdas2000analysis} assessed the risk of suicide by detecting the fluctuations in the fundamental frequency of people's speech. 
Dibeklio{\u{g}}lu \textit{et al.} \cite{dibekliouglu2015multimodal} explored the use of vocal prosody for depression detection.
Similarly, Syed \textit{et al.} \cite{syed2017depression} investigated the use of turbulence in speech patterns.

Besides, in AVEC challenges, low-level descriptors of voice signals, such as Mel-frequency Cepstral Coefficients (MFCCs), are provided, leading to many multi-modal methods incorporating these acoustic features for distress and mental illness detection \cite{yang2017multimodal, xing2018multi}.

\subsection{Body modality}

A few previous studies attempted to include the body modality in their models to predict psychological distress, mostly by extracting generic features from the video recordings related to the body.
For example, Joshi \textit{et al.} \cite{joshi2013can} computed Histogram of Gradients (HOGs) and Histogram of Optical Flow (HOFs) around the generic Space-Time Interest Points (STIPs) extracted from the videos, and then generated a ``Bag of Body Dynamics'' feature that was used for depression classification. Some of the multi-modal work presented in the AVEC challenges\cite{yang2018bipolar, xing2018multi, ray2019multi, yin2019multi} utilize the low-level descriptors of visual signals (such as latent CNN layer activation of ResNet\cite{he2016deep} and VGGNet\cite{simonyan2014very}) to predict on psychological distress.

More recent works also investigate the specific movement of body parts.
In the past few years, the skeletal models, either using RGB such as OpenPose \cite{cao2018openpose} or RGBD such as Microsoft Kinect SDK skeleton tracker\footnote{https://developer.microsoft.com/en-us/windows/kinect/}, have gained popularity for action recognition tasks and were used to generate more specific and concrete features by feature engineering \cite{mahmoud2013automatic, jaiswal2017automatic}.
For example,  Jaiswal \textit{et al.} \cite{jaiswal2017automatic} extracted head movements using Kinect and performed multi-modal classification with other audiovisual features to predict ADHD and ASD.
Though promising, the related work using such skeletal models on detecting psychological distress is still sparse.



In terms of automatic detection of self-adaptors, the only previous work that attempted to detect fidgeting behavior was presented by Mahmoud \textit{et al.} \cite{mahmoud2013automatic}. They developed a multi-modal framework for automatic detection of descriptors of rhythmic body movement by extracting Speeded-Up Robust Features (SURFs) interest points around Microsoft Kinect pose points and then detected rhythmic behaviors from analyzing the trajectories of the interest points. 
However, there are two limitations in their proposed automated system when applied to distress detection: 
1) Their dataset they used was based on acted data, so the behavior detected is not natural. For example, in more real interview scenarios, participants do not always fidget with a rhythmic pattern.
2) The trajectory data was noisy, and their method could not sufficiently handle the complexity of the detected body signal. As such, they were only able to achieve 59\% recognition on their acted dataset.

\subsection{Multi-modal Learning}

Since psychological distress is expressed through all modalities, many of the state-of-the-art models that predict signs of psychological distress proposed multi-modal approaches \cite{yang2017multimodal, xing2018multi, ringeval2018avec, ringeval2017avec, yang2018bipolar, xing2018multi, ray2019multi, yin2019multi}, combining low-level features extracted from the face, speech, and text, which are usually the features publicly available for the datasets. By only working with extracted features, most of these works focused on exploiting the given features, instead of analyzing the behavioral cues (e.g., specific gestures) of psychological distress.
For example, the winner of AVEC 2019 \cite{ray2019multi} proposed multi-layer attention fusion frameworks, but they did not explore the psychological basis of their models' decisions due to the lack of access to the raw data. 


\section{Dataset}\label{sec:dataset}


In this section, we describe the data collection, experimental design, and general characteristics of our collected dataset. This dataset is designed to enable investigation of the body modality for use in automatic detection of distress. 
 
Currently, the corpus is not publicly available due to the sensitivity of the collected video. Longer-term, we intend to make some portions (such as the features) of the data more broadly available to the research community.

\subsection{Overview and design}
Participants were recruited through the University of Cambridge email lists, student social media groups, and paper fliers posted around the town. 
We aimed to balance the sample with regards to distress levels, such that the database includes participants at the two distinct ends of the distress spectrum.
To identify participants with high versus low levels of distress, we conducted an online screening with a total 106 people who signed up for the study.
Participants completed standardized measures of depression (
PHQ-8~\cite{Kroenke:2009ky, Kroenke:2001fl} and anxiety
(
GAD-7 \cite{Spitzer:2006gb}), as well as demographics. 
In the selection, we balanced the participants according to the public norm shown in Table \ref{table:dataset-published-norms} (e.g., For depression, above 6.63 is marked as high, otherwise low).
Given potential gender differences in nonverbal communication \cite{mehrabian1972nonverbal}, we also balanced the final sample with regards to gender within each distress group\footnote{
Non-binary/other was given as an option in the registration form.
A number of people registered with this option.
However, none of those people met the distress level criteria and were thus not selected for an interview.}.
 From the initial screening, 35 were invited to the face to face session, including 18 with high distress and 17 with low distress. 

The participants completed the same measures of depression and anxiety immediately before the interview. This was meant to provide an assessment of distress closer in time to the interview and to increase the psychological salience of this information during the interview.

We adopted a data collection methodology inspired by the DAIC dataset collection method~\cite{Gratch:2014vo}, which consists of a human interviewer asking a series of open-ended conversational questions to elicit naturalistic behavior. The interviews were performed by a computer science researcher based on peer-support interview questions collected from the university support services. 
To achieve the conversational interview dynamic the interviewer asks general questions regarding the
participant's life and further encourages the participant to elaborate.
For example, the interviewer would ask \enquote{can you tell me about one time in your life you were particularly
happy?\@} and then ask some follow up questions regarding the example the participant provided.
The interviewer was blind to the distress level of participants during the interview.

To keep behaviors naturalistic, participants were not aware of the main goal of the study, which is an automatic analysis of behavioral cues. 
Instead, they were told that the experiment aimed at building models that can help in mental well-being. This ensured that their behavior would be as natural as possible. All participants got debriefed of the main aim of the data collection at the end of the session.  Participants were not informed of the results of their questionnaires, and all of them were handed a small booklet with the list of peer support and mental well-being services provided by the university. It is worth mentioning that the interviewer was blind to whether participants were from high or low distress groups in order not to affect their behavior. 
They were also instructed to limit their body and facial expressions throughout the interview and keep their sitting posture constant through all the interviews in order to avoid any changes in participants' behavior due to mimicry effect \cite{hess1999mimicry}. 

The dataset is labeled with participant responses to self-evaluation questionnaires right before the interview for assessing distress and personality traits, as well as demographic labels such as gender.
The distress questionnaires include the PHQ-8 for depression,
 GAD-7 for anxiety, SSS-8~\cite{Gierk:2014fv} for somatic symptoms,
 and the PSS~\cite{PerceivedStressSca:1983dv} for perceived stress.
 Personality traits are measured using the Big Five Inventory~\cite{John:ud}.
In sum, each participant provided responses to 5 questionnaires, in which PHQ-8 and GAD-7 were measured twice, both at registration and before the face-to-face session.

As a result, the dataset includes videos of fully natural non-acted expressions, including facial expressions, body motion, gestures, and speech. 

 \subsection{Preliminary Analysis}\label{subsec:dataset-initial-analysis}

 We collected videos of 35 interviewed participants with a total video duration of 07:50:08.
 General statistics regarding the questionnaire and demographic results within the dataset are provided in Table~\ref{table:dataset-primary-statistics}.
 Covariance is presented as normalized covariance values, also known as the correlation coefficient.
\subsubsection*{Confounding Correlations}
 
 \begin{table}[t]
  \small
  \begin{center}
   \begin{tabular}{rrrr}
    \toprule
    \textbf{Label} &
    \textbf{Range} &
    \textbf{Mean} &
    \makecell{\textbf{Covariance}\\\textbf{with Depression}}
    \\ [0.5ex]

    \midrule
    \makecell[l]{\textbf{Distress}} &&&  \\
    Depression & 0--19 & 7.43 & - \\
    Anxiety & 0--19 &  7.00 & 86.15\% \\
    Perceived stress & 1--30 & 18.17 & 84.00\% \\
    Somatic symptoms & 1--27 & 9.06 & 74.16\% \\
    \midrule
    \makecell[l]{\textbf{Personality}} &&&  \\
    Extraversion & 3--31 & 16.37 & -30.49\% \\
    Agreeableness & 12--34 & 25.67 & -42.21\% \\
    Openness & 7--39 & 27.29 & 4.29\% \\
    Neuroticism & 1--31 & 16.86 & 80.00\% \\
    Conscientiousness & 10--36 & 21.46 & -46.41\% \\
    \midrule
    \makecell[l]{\textbf{Demographic}} &&&  \\
    Gender & \multicolumn{2}{r}{18 M \& 17 F} & 9.47\% \\
    Age & 18--52 & 25.40 & -11.09\% \\

    \bottomrule
   \end{tabular}
  \end{center}
  \caption{
  \small
  General statistics regarding the questionnaire and demographic results within the dataset.
  This table demonstrates there are no confounding correlations with the depression label.
  }
  \label{table:dataset-primary-statistics}
 \end{table}

 We assessed confounding correlations based on the depression label, as much of the related work focuses on depression.
 While the distress measures, anxiety, perceived stress, and somatic stress, were found to be strongly correlated with depression,
 the personality measures have below 50\% covariance with the exception of neuroticism, which is a trait characterized by negative emotionality, with 
 an 80\% covariance.
 The demographic measures, gender, and age were negligibly correlated, with 9.47\% and
 -11.09\% covariance, respectively. 
Finally, the interview duration was found to be not correlated with any questionnaire result (less than 25\% covariance with all labels). Thus, we can be confident that there are no confounding correlations with personality scores or demographics.
 

\subsubsection*{Published Norms}
A comparison of the mean values for distress and personality measures between our dataset and the
published norms is presented in Table~\ref{table:dataset-published-norms}.
While there are differences, the measures are generally in line with the published norms. 
The dataset has a substantially higher mean perceived stress
score, but only slightly higher mean scores for anxiety and depression. 
Depression, extraversion, and neuroticism measures are particularly close to their published norms.
While the dataset mean for agreeableness and openness are substantially higher than the published
norms (over 10\% over the technical range for those measures). 
\begin{table}[h]
  \small
  \begin{center}
    \begin{tabular}{rrrr}
      \toprule
      \textbf{Label} &
      \textbf{Mean} &
      \textbf{Norm} &
      \textbf{Source}
      \\ [0.5ex]

      \midrule
      \makecell[l]{\textbf{Distress}} &&&  \\
      Depression & 7.43 & 6.63 & Ory et al.\ \cite{Ory:2013bs} \\
      Anxiety & 7.00 & 5.57 & Spitzer et al.\ \cite{Spitzer:2006gb} \\
      Perceived stress & 18.17 & 12.76 & Cohen et al.\ \cite{PerceivedStressSca:1983dv} \\
      Somatic symptoms & 9.06 & 12.92 & Gierk et al.\ \cite{Gierk:2014fv} \\
      \midrule
      \makecell[l]{\textbf{Personality}} &&& \\
      Extraversion & 16.37 & 16.36 & Srivastava et al.\ \cite{Srivastava:tc} \\
      Agreeableness & 25.67 & 18.64 & Srivastava et al.\ \cite{Srivastava:tc} \\
      Openness & 27.29 & 19.61 & Srivastava et al.\ \cite{Srivastava:tc} \\
      Neuroticism & 16.86 & 16.08 & Srivastava et al.\ \cite{Srivastava:tc} \\
      Conscientiousness & 21.46 & 18.14 & Srivastava et al.\ \cite{Srivastava:tc} \\

      \bottomrule
    \end{tabular}
  \end{center}
  \caption{
  Comparison of the mean questionnaire values within our dataset to the published norms.
  This shows that the population distribution, with regards to these distress and personality measures,
  is generally in line with the broader population.
  }
  \label{table:dataset-published-norms}
\end{table}



\subsection{Remarks}\label{subsec:difference-from-registration}
Participants completed the PHQ-8 and GAD-7 questionnaires twice: during registration and with the interview
process.
These questionnaires are temporal; specifically, they relate to the participant's mental state in
the past two weeks.
Given this, some difference between registration and interview results was expected.

With the exception of a small number of outliers, participants were generally consistent
in self-evaluation between registration and interview.
PHQ-8 responses had a mean difference of 0.89, while GAD-7 responses had a mean difference of 0.63.
As a result, we took the most recent response to self-evaluation questionnaires as the label for each participant's video recording.


\section{Method\label{sec:method}}
We used our collected dataset to study body gestures and self-adaptors. In this section, we demonstrate two different methods to analyze the body modality within the context of psychological distress. As a first step, we extract the most common audio-visual features. Then we describe a set of generic statistical body features that we extract to analyze general body gesture movement. To look specifically for self-adaptors, we then present an automatic approach to extract self-adaptors and fidgeting behavior in our dataset. We then perform a feature-based statistical analysis on the extracted body features - both generic and fidgeting features to understand what features are generally correlated with distress classification. Lastly, we move on to propose a multimodal approach to demonstrate further the effectiveness of body modality, where we incorporate and analyze the co-occurrence of multiple modalities to make predictions.

\subsection{Audiovisual Feature Extraction}\label{sec:meta_feature_extraction}
\subsubsection{Visual Features}
\label{sec:visual_features}
For each video, we used state-of-the-art tools, OpenPose \cite{cao2018openpose} and OpenFace 2.2 \cite{baltrusaitis2018openface}, to extract body pose features, facial Action Units (AUs), and gaze directions.

However, OpenPose and OpenFace do not take into account the consistency of the keypoints across time, causing the keypoints to usually fluctuate highly in many parts, introducing noise to the real continuous face and body motion. 
Besides, there are some frames where OpenPose or OpenFace fail to extract all pose points or gaze features, respectively.
To overcome these problems, we infer the missing data via Cubic Spline Interpolation across the whole sequence.
We then smooth the data using a Savitzky-Golay filter \cite{schafer2011savitzky} (window length is 11 and the order of the polynomial is 3). 

\subsubsection{Audio Features}
Speaker diarization involves partitioning an audio stream into homogeneous segments according to the speaker's identity. In order to distinguish the speech of the interviewer and the participant, we use the open-source Speaker-Diarization project~\cite{speakerdiarization} which utilizes an Unbounded Interleaved-State Recurrent Neural Network (UIS-RNN) \cite{zhang2019fully}, to extract speaker identities with respect to the time axis. 
We then conduct a manual check to assign correct diarization labels to the participant and the interviewer.
We also use pyAudioAnalysis \cite{giannakopoulos2015pyaudioanalysis} to extract MFCCs.

\subsection{Generic Body Features} \label{sec:meta_gesture_feature_extraction}

To explore the body modality, we extract and analyze the set of generic statistical features that describe the body movements.
\subsubsection{Feature Extraction}
Two kinds of statistical features are computed and extracted: global features and localized features.
In the global features, we care about the overall statistics of motion, while in the localized features (features that are within specific body parts, such as head, hands, and legs), we are interested in the statistics of the motion within the body parts, which we refer to as ``localization''. Our notation is summarized in Table \ref{table:results-feature-notation-mapping}.

We define a ``gesture'' as a period of sustained movement within a body localization. For example, waving hands is a gesture within ``\texttt{Hn} (hand)'' localization, and shaking legs continuously will register a gesture in ``\texttt{L} (Legs)'' localization.

To detect gestures within a localization, we scan the video using a moving window method.

First, the per-frame absolute movement ($L^2$ distance) is calculated for each pose point.
The value is then averaged by the number of pose points in the localization.
Formally,
\begin{equation}
  \begin{split}
    F_{t} & = \dfrac{1}{|P|}\sum_{p \in P} ||P_{p,t} - P_{p,t-1}||_{2}
  \end{split}
  \label{eq:gesture-detection-1}
\end{equation}
where $P_{p, t}$ is the position
vector of pose point $p$ at time $t$, and $F_{t}$ is the averaged per-frame movement across all points. $P$ are the collection of pose points in this localization.

Second, a moving window is applied such that a small number of frames do not have a
disproportionate effect on the detection. This process can be expressed by:
\begin{equation}
  \begin{split}
    W_{i} = \dfrac{1}{l}\sum_{t = i \times l}^{t < i \times (l + 1)}F_{t}
  \end{split}
\end{equation}
where $W_{i}$ is the windowed average at window index $i$, $l$ is the length of the window, and
$F_{t}$ is the average movement at frame $t$, from Equation~\ref{eq:gesture-detection-1}.
We experimentally chose $l = 10$, i.e.\ a second of movement is represented by $3$ windows.

Third, the window moves until an average movement
above a threshold is found, which is considered the beginning of the gesture.
The gesture continues until $n=3$ consecutive windows (30 frames, approximately 1 sec) are found below the movement threshold, which is thus considered the end of the gesture.

Table \ref{table:results-feature-notation-mapping} lists the set of body features we extract. Below we explain how we define each of these features for the overall body. Similarly, the localized features can be calculated for every localization/body part.
\begin{itemize}
  \item \textbf{Average frame movement} - the per-frame average movement (moving distance) of every pose point of the body. This is the only feature that is not based on detected gestures.
  \item \textbf{Proportion of total movement occurring during a gesture} - the proportion of total
  movement that occurred while a gesture is happening (within some localizations).
  \item \textbf{Average gesture surprise} - 
  defined as 
  ``fraction of frames with no gesture happening'' $\div$ ``number of gestures''.
  
  
  For example, if two gestures occurred within a sample such that 80\% of the sample
  duration had no gesture occurring, the average gesture surprise would be $\dfrac{80\%}{2}=40\%$.
  Whereas, if there were 100 gestures, 
  the average surprise is 0.8\%, even though both samples had the same proportion without
  any gesture occurring.
  This matches the intuition that each gesture within 100 evenly spaced gestures would be unsurprising
  as they were regularly occurring, whereas the 2 evenly spaced gestures would be surprising because
  nothing was happening in between.
  \item \textbf{Average gesture movement standard deviation} - the standard deviation of per-frame
  movement within a gesture is averaged across all detected gestures.
  This is intended to indicate the consistency of movement intensity
  through a gesture.
  \item \textbf{Number of gestures} - total number of detected gestures across all tracked localizations.


\end{itemize}
\begin{table}
  \begin{minipage}{.35\linewidth}
    \vspace{-0.7cm}
    \begin{center}
      \begin{tabular}{cc}
        \toprule
        \textbf{Localization} & \textbf{Abbr.} \\ [0.4ex]
        \midrule
         \textbf{\underline O}verall & \texttt{O} \\
        \cmidrule{1-2}
         \textbf{\underline H}a\textbf{\underline n}ds & \texttt{Hn} \\
        \cmidrule{1-2}
         \textbf{\underline{ He}}ad & \texttt{He} \\
        \cmidrule{1-2}
        \textbf{\underline L}egs & \texttt{L} \\
        \bottomrule
      \end{tabular}
    \end{center}
  \end{minipage}
  \begin{minipage}{.5\linewidth}
    \begin{center}
      \begin{tabular}{ccc}
        \toprule
        & \textbf{Feature} & \textbf{Abbr.} \\ [0.5ex]
        \midrule
        \parbox[c]{3mm}{\multirow{5}{*}{\rotatebox[origin=c]{90}{\textbf{Overall\ \ \ \ \ \ \ \ \ \ \ \ }}}}
        & Average  \textbf{\underline{F}}rame  \textbf{\underline{ M}}ovement & \texttt{FM} \\ \cmidrule{2-3}
        & \makecell{Proportion of total  \textbf{\underline M}ovement \\ occurring during a  \textbf{\underline G}esture} & \texttt{GM} \\ \cmidrule{2-3}
        & Average  \textbf{\underline G}esture  \textbf{\underline S}urprise & \texttt{GS} \\ \cmidrule{2-3}
        & \makecell{Average  \textbf{\underline G}esture movement \\ standard  \textbf{\underline D}eviation} & \texttt{GD} \\ \cmidrule{2-3}
        & \textbf{\underline N}umber of  \textbf{\underline G}estures & \texttt{GN} \\
        \midrule
        \parbox[c]{3mm}{\multirow{5}{*}{\rotatebox[origin=c]{90}{\textbf{Localized\ \ \ \ \ \ }}}}
        & Average  \textbf{\underline L}ength of  \textbf{\underline G}esture & \texttt{GL} \\ \cmidrule{2-3}
        & \makecell{\textbf{\underline{A}}verage per-frame \\ \textbf{\underline G}esture movement} & \texttt{GA} \\ \cmidrule{2-3}
        &  \textbf{\underline T}otal movement in  \textbf{\underline G}estures & \texttt{GT} \\ \cmidrule{2-3}
        & Average  \textbf{\underline G}esture  \textbf{\underline S}urprise & \texttt{GS} \\ \cmidrule{2-3}
        & \textbf{\underline{N}}umber of  \textbf{\underline G}estures & \texttt{GN} \\
        \bottomrule
      \end{tabular}
    \end{center}
  \end{minipage}
  \\
  \caption{Feature notation Abbrs. of \texttt{BodyGesture}.}
  \label{table:results-feature-notation-mapping}
\end{table}

\subsubsection{Feature Processing}
All the movement data is extracted from smoothed OpenPose data described in Section \ref{sec:visual_features}. All these body gesture features are concatenated (thereafter marked as \texttt{BodyGesture}, which has a feature vector of length 20 for each participant) and all features are normalized such that the length of the sample does not affect the results.

Sum-based features (e.g., \ gesture length, gesture count, total movement, etc.) are normalized against
the total number of frames in the sample.
Gesture average features, such as gesture surprise, are again normalized against the total number of gestures.

\subsection{Self-adaptors and Fidgeting features}\label{sec:automatic_fidget_detection}
In addition to the generic body features, we were interested in analyzing the self-adaptors and fidgeting behavior. In this section, we present our fidgeting detection system in three subsections. We start by exploring the self-adaptors/fidgeting encoding and the overall hierarchical design. Then we show the methods of building the two essential detectors of our hierarchical model in the following two subsections. For each detector, we demonstrate the detector's design, and then present the labeling strategy which provides reliable labels for training and evaluation. In order to validate the effectiveness of our automated fidgeting detection approach before moving onto distress classification, we evaluate our model thoroughly both on an acted dataset and on our newly collected dataset of natural expressions.

\subsubsection{Overall Design and Encoding}
\begin{figure}[h]
    \centering
    \includegraphics[width=9cm]{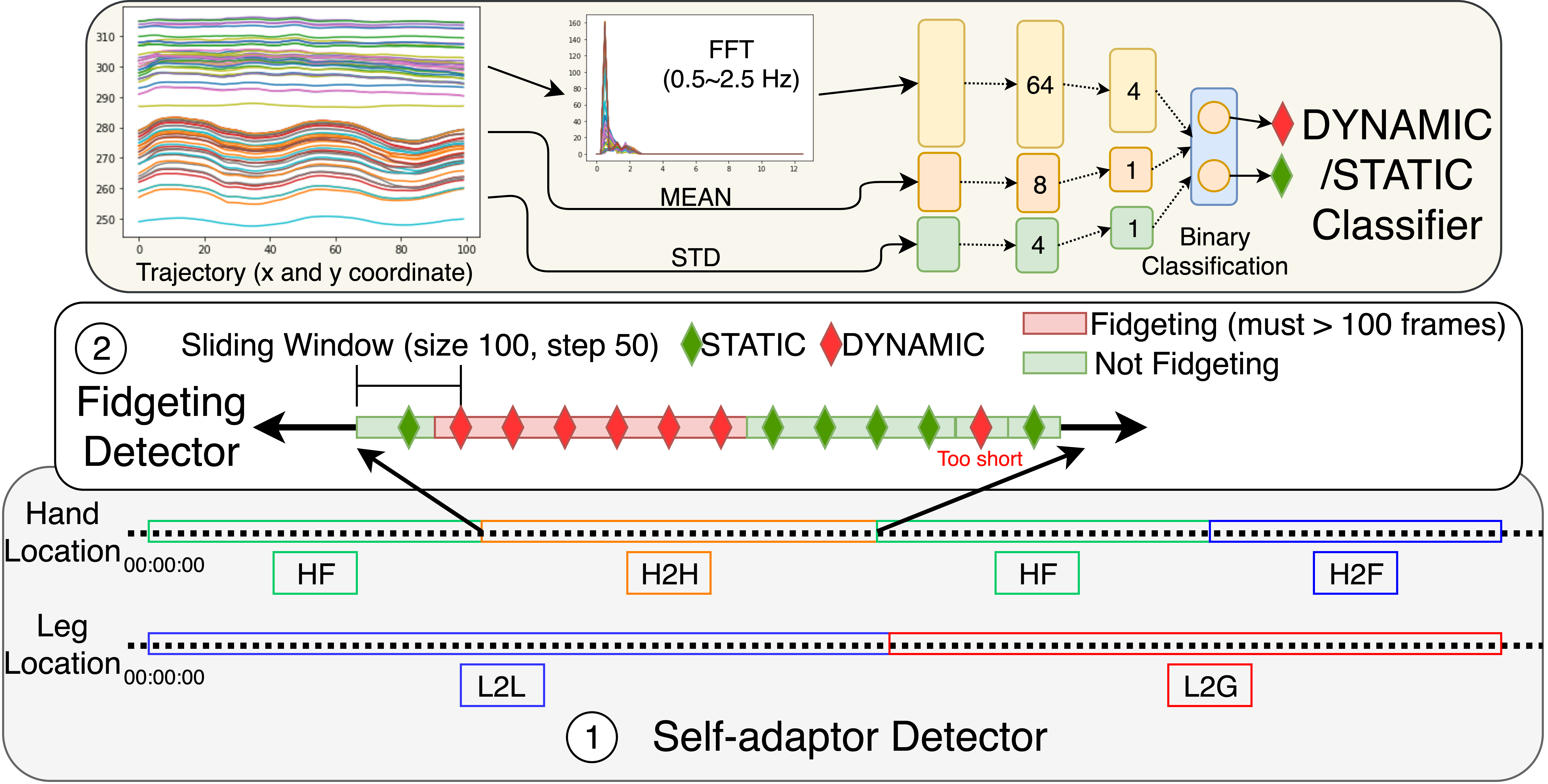}
    \caption{Hierarchical self-adaptor detection workflow. (1) First, detect hand/leg location; (2) Classify motion using \textit{DYNAMIC/STATIC Classifier} and then finally combine location and motion to give high-level fidgeting event. The figure shows the detection of \texttt{H2H} (Hand to hand) fidget. The same principle applies to other fidgets.}
    \label{fig:fidgeting_detection_workflow}
\end{figure}
\begin{table}[]
    \centering
    \begin{tabular}{lp{4.5cm}}
        \toprule
        \textbf{Self-adaptors} & \textbf{Description} \\
        \midrule
        \texttt{H2H} & \textbf{\underline H}and to \textbf{\underline H}and \\
        \texttt{H2A} & \textbf{\underline H}and to \textbf{\underline A}rm \\
        \texttt{H2L} & \textbf{\underline H}and to \textbf{\underline L}eg \\
        \texttt{H2F} & \textbf{\underline H}and to \textbf{\underline F}ace \\
        \texttt{HF} & \textbf{\underline H}and \textbf{\underline F}ree (when not belong to any of above) \\
        \texttt{L2G} & Both \textbf{\underline L}egs on \textbf{\underline G}round \\
        \texttt{L2L} & \textbf{\underline L}eg on the other \textbf{\underline L}eg (crossed legs) \\
        \midrule
        \textbf{Action Events} & \textbf{Description}\\
        \midrule
        DYNAMIC & Moving obviously \\
        STATIC & No obvious movement is observed \\
        \bottomrule
        \\
        \toprule
        \textbf{Fidgeting Type} & \textbf{Combination} \\
        \midrule
        \texttt{CHF} (\textbf{\underline C}ross \textbf{\underline H}and \textbf{\underline F}idgeting) & \texttt{H2H} + DYNAMIC \\
        \midrule
        \texttt{SHF} (\textbf{\underline S}ingle \textbf{\underline H}and \textbf{\underline F}idgeting) & \{\texttt{H2A}, \texttt{H2L}, \texttt{H2F}, \texttt{H2F}\} + DYNAMIC \\
        \texttt{SHF-L} (to \textbf{\underline L}eg only) & \texttt{H2L} + DYNAMIC \\
        \texttt{SHF-F} (to \textbf{\underline F}ace only) & \texttt{H2F} + DYNAMIC \\
        \texttt{SHF-A} (to \textbf{\underline A}rm only) & \texttt{H2A} + DYNAMIC \\
        \midrule
        \texttt{LFF} (\textbf{\underline L}eg/\textbf{\underline F}eet \textbf{\underline F}idgeting) & \{\texttt{L2G}, \texttt{L2L}\} + DYNAMIC\\
        \bottomrule
    \end{tabular}
    \vspace{0.1cm}
    \caption{Self-adaptor and fidgeting encoding book}
    \vspace{-0.6cm}
    \label{tab:event_list}
\end{table}

Given the lack of broad agreement on the definition of fidgeting so far, we utilize a two-step hierarchical model to identify fidgeting.
As shown in Table \ref{tab:event_list}, we first identify self-adaptors, which we define as low-level location events (e.g.\ \texttt{H2H}, \texttt{H2F}). 
Secondly, action events (i.e.\ DYNAMIC, STATIC) of hand/leg are classified by the \textit{DYNAMIC/STATIC Classifier}. Fidgeting is then defined as a combination of low-level self-adaptors and action events.
Specifically, we define three types of fidgeting: cross hand fidgeting, single-hand fidgeting, and leg/feet fidgeting.



\subsubsection{Self-adaptor Detector}

\paragraph{Design}
Each body location is represented using a bounding box. Self-adaptors are defined as overlapping bounding boxes.
We represent the hand and face using the smallest rectangular box bounding all corresponding hand or face keypoints.
The forearms, upper arms, lower legs, and upper legs`bounding boxes' long sides are aligned with the connection between two joints from OpenPose, while the width is a free parameter tuned for the best automatic detection performance.

First, \texttt{H2H} self-adaptor events are detected (i.e.,\ when the two hands' bounding boxes overlap). 
Then all other hand-based self-adaptor events are detected, for all segments of the video not containing \texttt{H2H} segments. 

All self-adaptors, except for \texttt{H2F}, must be longer than 100 frames (around 4 seconds with the frame rate of 26). 
This reduces noise from detected self-adaptor events.



\paragraph{Labeling and Evaluation}
In order to validate our self-adaptor detector, we manually labeled 4 participants' videos, a total duration of 59 minutes. 
The inter-labeler agreement was checked using Krippendorff's alpha.
Each frame was labeled with one of the self-adaptor codes from Table \ref{tab:event_list}. 
Within these videos, participants perform different self-adaptors and each event has a minimum total duration of 5 minutes, with the exception of \texttt{H2F}. 

As shown in Table \ref{tab:location_detector_evaluation}, the  Krippendorff’s alpha agreement for left-hand location is 0.823, for right-hand location is 0.888 and for leg location is 1.00. 
This suggests good agreement between the annotators and, thus, the reliability of the labels. 
The results show that our network is able to detect self-adaptor with excellent overall precision, and especially for the \texttt{H2H}, \texttt{H2F}, \texttt{L2L} and \texttt{L2G} events, the detector reached a very high accuracy respectively. 
Note that, `NA' in Table \ref{tab:location_detector_evaluation} means that there is no corresponding gestures in the evaluation set of 4 labelled participants.


\begin{table}[!htbp]
\centering
\begin{tabular}{lccc}
\multicolumn{4}{l}{Hand Self-adaptors (left/right)}\\
\toprule
 & Precision & Recall & F1 Score \\
\midrule
\texttt{H2H} & 1.00/1.00 & 0.99/0.99 & 1.00/1.00 \\
\texttt{H2A} & 1.00/NA & 0.64/NA & 0.79/NA \\
\texttt{H2L} & 0.96/0.88 & 0.86/0.82 & 0.91/0.85 \\
\texttt{H2F} & NA/1.00 & NA/1.00 & NA/1.00 \\
\texttt{H2F} & 0.63/0.83 & 0.99/0.98 & 0.77/0.90\\
\midrule
Alpha Score: & 0.823/0.888 \\
\bottomrule
\\
\multicolumn{4}{l}{Leg Location}\\
\toprule
{} & Precision & Recall & F1 Score \\
\midrule
\texttt{L2L} & 1.00 & 1.00 & 1.00\\
\texttt{L2G} & 1.00 & 1.00 & 1.00 \\
\midrule
Alpha Score: & 1.000 \\
\bottomrule
\end{tabular}
\vspace{0.2cm}
\caption{Self-adaptor Detection Evaluation}
\vspace{-5mm}
\label{tab:location_detector_evaluation}
\end{table}

\subsubsection{Fidgeting Detector} \label{sec:Fidgeting_Detector}
\paragraph{Design}
As shown in Fig. \ref{fig:fidgeting_detection_workflow}, the \textit{DYNAMIC/STATIC Classifier} operates on extracted \textbf{optical flow} from a sliding window across the video (size 100 frames, step 50 frames).
To classify the action (DYNAMIC/STATIC), hand movements (especially fingers) and leg movements require optical flow to obtain smooth trajectories, given OpenPose estimations become unreliable when hands intersect or are occluded. We thus initialize the optical flow with the OpenPose estimations at the beginning of each slice.

We choose Fast Fourier Transform (FFT), standard deviation (STD), and mean values (MEAN) of point trajectories as our input features (in this case, number of trajectories is 2 $\times$ number of keypoints as we have 2-D data for each keypoint). 
For fidgeting, we are more interested in the cyclic motion with a frequency ranging from 0.5Hz to 2.5Hz \cite{mahmoud2013automatic}. Therefore, we extracted the spectrum data within the range $[0.5, 2.5]$ Hz.
As we analyze slices of length 100, the dimension of FFT spectrum data that is within $[0.5, 2.5]$ Hz is always fixed at $41 \times$ number of trajectories. 
We then average over the FFT values that have the same frequency to produce an FFT feature of length 41. As for the STD and MEAN features, we simply calculate along the time axis and give a vector with a length of the number of trajectories for each feature.

\paragraph{Labeling and Evaluation}
To train and evaluate the \textit{DYNAMIC/STATIC Classifiers}, accurate labeling is required. Three classifiers are required to cover the three categories of detected self-adaptors: \{\texttt{H2H}\}, \{\texttt{H2A}, \texttt{H2L}, \texttt{H2F}, \texttt{H2F}\}, and \{\texttt{L2G}, \texttt{L2L}\}.

We labelled DYNAMIC/STATIC on each of the three categories. We randomly sampled and labeled approximately 30\% of slices for each category in every video. 

Two researchers labeled the data independently. As shown in Table \ref{tab:action_detector_labelling}, we first manually dropped the slices with a wrong category label (e.g.\ a slice is detected as \texttt{H2H} while it's in fact not). The number of slices that have a correct category label is shown as ``Correct''. Secondly, we labeled DYNAMIC/STATIC and dropped the slices that lack a consensus between two researchers. The number of slices with an agreement is shown as ``Agreed''. The high percentage of both ``Correct'' and ``Agreed'' suggests the good performance of our self-adaptor detection and also the high reliability of action labels.




\begin{table}[htp]
    \centering
    \begin{tabular}{lcccc}
        \toprule
        Category & Total & Correct & Agreed \\
        \midrule
        BOTH: \texttt{H2H} & 3962 & 3922 (99\%) & 3793 (96\%)\\
        LEFT:\{\texttt{H2A}, \texttt{H2L}, \texttt{H2F}, \texttt{H2F}\} & 1614 & 1566 (97\%) & 1539 (96\%)\\
        RIGHT:\{\texttt{H2A}, \texttt{H2L}, \texttt{H2F}, \texttt{H2F}\} & 1620 & 1588 (98\%) & 1563 (96\%)\\
        \{\texttt{L2G}, \texttt{L2L}\} & 6536 & 6536 (100\%) & 6196 (95\%)\\
        \bottomrule
    \end{tabular}
    \vspace{0.1cm}
    \caption{Hand/Leg action labelling overview}
    \vspace{-5mm}
    \label{tab:action_detector_labelling}
\end{table}

Having reliable slice labels, we then partitioned participants into 5 folds and performed slice-level cross-validation.
For evaluation, we calculated accuracy, F1 score, and their respective standard deviations.

\begin{table}[htp]
    \centering
    \begin{tabular}{lcccc}
        \toprule
        Category & Acc. & Acc. Std. & F1 & F1 Std.\\
        \midrule
        BOTH: \texttt{H2H} & 0.833 & 0.019 & 0.834 & 0.019\\
        LEFT:\{\texttt{H2A}, \texttt{H2L}, \texttt{H2F}, \texttt{H2F}\} & 0.884 & 0.025 & 0.884 & 0.026\\
        RIGHT:\{\texttt{H2A}, \texttt{H2L}, \texttt{H2F}, \texttt{H2F}\} & 0.895 & 0.026 & 0.894 & 0.026\\
        \{\texttt{L2G}, \texttt{L2L}\} & 0.875 & 0.022 & 0.871 & 0.021\\
        \bottomrule
    \end{tabular}
    \vspace{0.1cm}
    \caption{\textit{DYNAMIC/STATIC Classifier} evaluation (LEFT means left hand, RIGHT means right hand, BOTH means both hands)}
    \vspace{-0.6cm}
    \label{tab:action_detector_evaluation}
\end{table}

As shown in Table \ref{tab:action_detector_evaluation}, the detector achieved generally high accuracy and F1 score with low standard deviations. Though the hand actions are difficult even for researchers to label, the detector can successfully classify more than 80\% of slices.
\vspace{-0.3cm}
\subsection{Feature encoding}\label{sec:feature_encoding}
This section describes how we encoded low-level frame-level features described in Sec \ref{sec:meta_feature_extraction} and \ref{sec:automatic_fidget_detection} in preparation for the final prediction step. The generic statistical \texttt{BodyGesture} will not need to be encoded since it represents global statistical features rather than time-series features.
\vspace{-0.3cm}
\subsubsection{Fidgeting features processing}
Having extracted low-level features from each frame, we combine them to form high-level descriptors of fidgeting behavior (\texttt{CHF}, \texttt{CHF}, and \texttt{LFF} as shown in Table \ref{tab:event_list}). 
The \texttt{Fidget\_pure} feature group is formed by \{\texttt{HCF}, \texttt{SHF-L}(left hand), \texttt{SHF-L}(right hand), \texttt{SHF-A}(left hand), \texttt{SHF-A}(right hand), \texttt{SHF-F}(left hand), \texttt{SHF-F}(right hand), \texttt{LFF}\}. 
The \texttt{Fidget\_pure} group is combined with a participant speaking feature array to form the full fidget feature group, enabling us to investigate whether fidgeting and speaking co-occurrence is relevant.
This participant speaking feature array indicates whether the participant is speaking during a frame.
This is calculated using the previously described diarization data.

After all the feature extraction, we have several feature groups shown in Table \ref{tab:feature_group}.
\begin{table}[]
    \centering
    \begin{tabular}{ccl}
    \toprule
        Feature Group & Dimension & Description\\
        \midrule
        \texttt{BodyGesture} & $20\times1$ & Body Gesture Statistical Features\\
        \texttt{Fidget} & $9\times N$ & Fidget feature \& Speaking array\\
        \texttt{Fidget\_pure} & $8\times N$ & Fidget feature only\\
        \texttt{Gaze} & $8\times N$ & Gaze direction \\
        \texttt{AUs} & 3$5\times N$ & Action Units \\
        \texttt{MFCCs} & $13\times N$ & \\
    \bottomrule
    \end{tabular}
    \vspace{0.2cm}
    \caption{Feature Groups. $N$ is number of frames in each recording of participants.}
    \label{tab:feature_group}
    \vspace{-10mm}
\end{table}
\vspace{-0.3cm}
\subsubsection{Per-frame representation}
\begin{figure*}[t]
    \begin{center}
        \includegraphics[width=15cm]{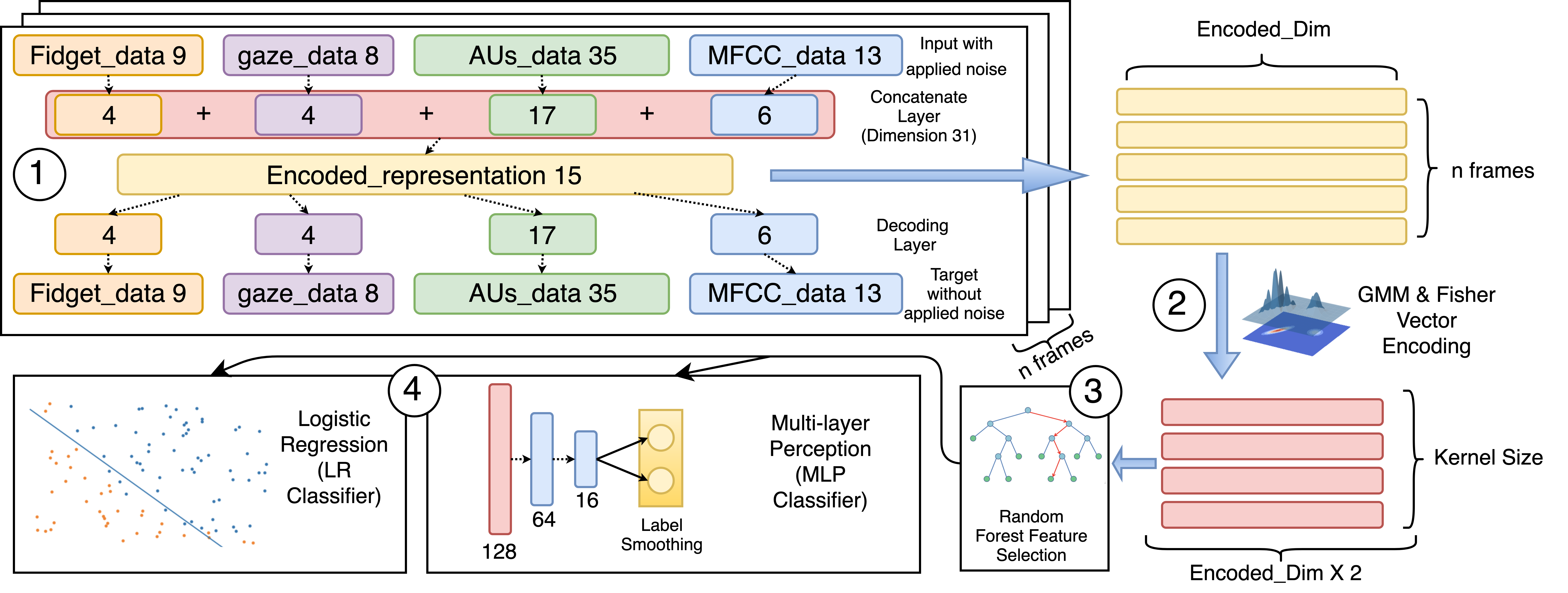}
    \end{center}
    \caption{
    Multi-modal fusion \& classification pipeline.
    The dashed arrow represents a fully connected neural network between dense layers.
    Pose estimation, gaze, Action Units, and MFCC data are extracted from videos.
    Fidget features are computed using the method described in Section \ref{sec:method}.
    (1) All features are fed into a Multi-modal Deep Denoising Auto-Encoder (multi-DDAE) to generate a compact per-frame encoded representation.
    (2) These per-frame features are then compressed into a whole video representation using a Gaussian Mixture Model (GMM) and Fisher Vector combination.
    (3) Random Forest feature selection is performed.
    (4) Finally, a classifier predicts a given label.
    We experiment with two classifiers, a logistic regression classifier and a Multi-layer Perception.}
    \label{fig:method_overview}
\end{figure*}
In order to capture more useful feature representations and reduce the dimensionality, and inspired by our previous work \cite{zhang2020multi}, different modalities are combined using a Multi-modal Deep Denoising Auto-Encoder (multi-DDAE). As shown in Fig.~\ref{fig:method_overview}, each modality is encoded through a dense layer and then all are concatenated to yield the last shared dense layer which provides the representation we use. The shared layer is then inversely decoded to generate each modality. We optimized the hyper-parameters of the auto-encoder via several experiments so that the dimensions of hidden layers are \{$0.5d, 0.25d, 0.5d$\} where $d$ represents the input dimension of each node, and the noise applied at the input is 0.1 Gaussian noise. The training optimization target is the joint Mean Square Error (MSE) of the MSEs of the feature group at each node (later we fixed the loss weights to be 0.35 for the fidget feature group while 0.1 for others, as we are more interested in fidgeting in our experiments).
\vspace{-0.2cm}
\subsubsection{Whole video representation}
Due to varying lengths of the videos, it's necessary to unify the dimensionality of the per-video representation. 
Though Fisher Vector was originally proposed to aggregate visual features \cite{perronnin2010improving}, it has become popular in social signal processing such as bipolar disorder \cite{syed2018automated} and depression recognition \cite{dhall2015temporally}.
Inspired by these applications, we apply a Gaussian Mixture Model to cluster similar per-frame representations and then use an Improved Fisher Vector encoding to obtain a fixed-length representation. As a result, the feature is transformed from $\texttt{num\_frames} \times \texttt{feature\_dim}$ to $2 \times \texttt{GMM\_Kernel\_num} \times \texttt{feature\_dim}$.

\vspace{-0.1cm}
\subsection{Classification of signs of distress} \label{sec:Distress_classification}
We apply a Random Forest to select important features from the per-video representation. The selected features are used by the classifier. We experiment with two classifiers: 1) a logistic regression-based classifier (LR) using a binary threshold of 0.5; 2) a Multi-Layer Perception (MLP) with two softmax outputs for binary classification.

As the available samples are limited and the useful features vary across individual differences, label smoothing~\cite{muller2019does} is applied to the MLP model in order to further boost the performance. 
More formally:
\begin{equation}
\begin{split}
    L\_new = L \times (1 - s) + \frac{s}{n}
\end{split}
\end{equation}
where $L$ is the one-hot label at softmax outputs, $s$ is the smoothing parameter, and $n$ is the number of classification classes.
For example, when smoothing is 0.2, the one-hot label \{0, 1\} will become \{0.1, 0.9\}, which lowers the confidence on training samples but reduces overfitting.

\section{Statistical Analysis of Body Gesture}\label{sec:metagesture_evaluation}
To better understand the effect of different body-related features, before moving to deep multimodal learning, we deploy a simple linear regression model to perform statistical analysis on the body gesture features (\texttt{BodyGesture} from Sec. \ref{sec:meta_gesture_feature_extraction}) and fidgeting features (\texttt{Fidget\_pure} from Sec. \ref{sec:feature_encoding}). The aim of this section is to shed some light on the effect of different movements of every part of the body and its correlation with depression. 
\vspace{-0.3cm}
\subsection{Experimental Setup}
Fidgeting features from \texttt{Fidget\_pure} is processed by averaging along the time axis ($9 \times N$ to $9 \times 1$) to match the dimension of other features in \texttt{BodyGesture} ($20\times1$).
Reporting notation is defined as \enquote{\texttt{[localization]-[feature type][linear polarity]}}. Localization and feature type token mappings are provided in Table~\ref{table:results-feature-notation-mapping}.
Polarity is defined below:
\begin{itemize}
    \item \enquote{\texttt{+}/\texttt{$\neg$}}: A greater value (e.g.\ more activity) contributing to a positive/negative classification
    \item \enquote{\texttt{/}}: A near-zero coefficient in linear model.
    \item \enquote{\texttt{?}}: The polarity is observed inconsistent in different folds of cross-validation.
\end{itemize}

\begin{table}[]
\centering
\begin{tabular}{ll}
\toprule
Feature Set                                                          & F1-score \\
\midrule
\texttt{O-FM} & 34.43\%  \\
\texttt{BodyGesture}  & 66.81\%  \\
Searched \texttt{BodyGesture} & 82.70\%  \\
\texttt{Fidget\_pure}  & 49.60\%  \\
Searched [\texttt{BodyGesture}, \texttt{Fidget\_pure}] & \textbf{83.38\%} \\
\bottomrule
\end{tabular}
\vspace{0.2cm}
\caption{Results of linear regression threshold classification on body gesture statistical features and fidget features. [A, B] represents a concatenation of feature vector A and B.}
\label{tab:feature_search}
\end{table}

With the linear model, we perform 3-fold cross-validation on depression labels, which is more reliable than normal train-valid-test split for our small dataset.
Cross-validation also provides more confidence about the polarity of each feature, as only the features that show consistent polarity across all folds will be marked.
All results are calculated as the mean of 3-fold cross-validation results.
All experiments and cross-validation are participant-independent.

\subsection{Results and Discussion}
As shown in Table \ref{tab:feature_search}, with only the global movement (\texttt{O-FM}), the F1 score is only 34.43\%. This means that measuring the quantity of global motion in the body is not enough indicator of depression.
While when combining all body gesture statistical features, the classifier achieves 66.81\% F1 score. 

Note that all body gesture statistical features include a large set of features representing statistics of different body parts as well as global body motion, as explained in Sec. \ref{sec:meta_gesture_feature_extraction}.
In order to filter out this large feature set, we performed an exhaustive feature search to obtain the combination of features that gives the best performance, repsresented in Table \ref{tab:feature_search} as ``Searched \texttt{BodyGesture}''.
It reaches a good F1 score at 82.70\%.

As shown in Table \ref{tab:feature_search}, when we combine specific fidgeting features (\texttt{Fidget\_pure}) with \texttt{BodyGesture}, and perform feature search on the concatenated feature,
the F1-score reaches the best at 83.38\%.
The resulted best feature combination includes: \{\texttt{O-FM?, O-GM+, O-GN?, Hn-GN?, Hn-GS$\neg$, He-GL+, He-GN+, He-GT+, He-GA+, He-GS+, L-GL+, L-GN+, L-GA+, SHF-L(Right)+, SHF-A(Right)+, SHF-F(Right)+, SHF-F(Left)+}\}. Looking deeply into this list of features we could infer some interesting insights into the overall body movements in our dataset, which we explain below.

For example, the \texttt{O-GM+} token suggests that more movement within gestures relative to all other movement is indicative of depression, and especially, total movement within head gestures (\texttt{He-GT+}) is positively correlated with depression.
The localized features suggest that the length of gestures in the head and legs (\texttt{He-GL+, L-GL+}) has a correlation with depression.
It's clear that gesture statistics in hands (\texttt{Hn-*}) are generally not interesting in prediction, while the classifier pays head and leg motions more attention.
However, \texttt{Hn-GS$\neg$} suggests that more regular (thus less surprising) hand gestures (e.g. constant fidgeting) show a positive contribution to depression.


We can also conclude that a higher quantity of right hand fidgeting on the leg, arm, and face (\texttt{SHF-*(Right)+}) have a positive contribution to the higher depression level, and left hand fidgeting on the face (\texttt{SHF-F(Left)+}) is also positively correlated with high depression level. 
The difference in left and right might be because most participants are right-handed and therefore, their left hands exhibit less useful motions that are predictive of depression.
The conclusion is not surprising, as, in our observations, people perform hand to hand fidgeting regardless of their depression label. 
Combining the results from above, we can conclude that, in our dataset, more regular hand gestures and more fidgeting on the leg, arm, and face are indicative of depression. Depressed participants also have more frequent motions in the head and leg region.

\section{Evaluation of Multimodal Deep Learning}\label{sec:results}

In this section, we evaluate and demonstrate the validity and potential of fidgeting features as complementing modality with other features to predict the signs of psychological distress.

First, we present some baseline distress classification results on our dataset.
Next, we present results for our full multi-modal classifier pipeline, where we investigate the effects of hyper-parameters on the performance given the small size of our dataset.
Finally, we apply our automatic fidgeting detection approach to a publicly available dataset \cite{mahmoud2013automatic} to demonstrate its accuracy and generalisability beyond our dataset.

As in Sec.~\ref{sec:metagesture_evaluation}, all results are calculated as the mean of 3-fold cross-validation results.
All experiments and cross-validation are participant-independent.
\vspace{-0.3cm}
\subsection{Baselines}
As a baseline, we used Gaussian kernel Support Vector Machines (SVMs) classifiers applied on each individual feature group used in our multi-modal model (listed in Table~\ref{tab:feature_group}).
Unlike in Sec.~\ref{sec:metagesture_evaluation}, non-linearity can be considered in these baseline models.
They are evaluated for a binary depression label and a binary anxiety label.
These models provide a simple and common baseline for our dataset.
For the baseline SVM, we use the mean value for each feature over the whole sample, thus providing a normalized representation with mean values of all the features.
Results are presented in Fig.~\ref{fig:final_results}.

These baseline models demonstrate two points: first, the behaviors we are attempting to classify in our dataset are complex; and second,
our fidgeting features by themselves are not trivially predictive of distress, but rather require learned representations. 





\vspace{-0.3cm}
\subsection{Multi-modal distress classification}
As presented in the previous baseline section, single modalities are not enough to capture the complexity of signs of psychological distress. 
Therefore we experiment with our proposed multi-modal classification framework.
We encode different modalities through multi-DDAE and Improved Fisher Vector encoding (Sec. \ref{sec:feature_encoding}), and classify distress labels using either LR or MLP classifier after Random Forest feature selection (Sec. \ref{sec:Distress_classification}).

In Fig.~\ref{fig:final_results}, we present the best performance of different feature group combinations using our multi-modal fusion framework. We use a Random Forests (RF) for feature selection. As RFs take in labels to find the most discriminative features, this feature selection is only performed on the training set and selected features are then applied to the test set, which prevents label leaking.


\begin{figure}
    \centering
    \includegraphics[width=7cm]{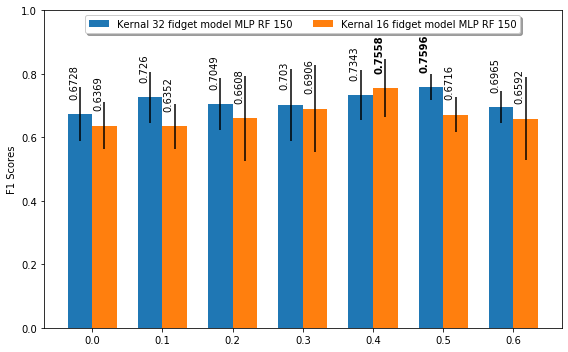}
    \caption{Effects of label smoothing. In general, smoothing can boost performance. (error bar extends by the standard deviation in either side and best performance in \textbf{bold})}
    \label{fig:label_smoothing}
\end{figure}

\subsubsection{Effects of some hyper-parameters}
\begin{figure}
    \centering
    \includegraphics[width=7.8cm]{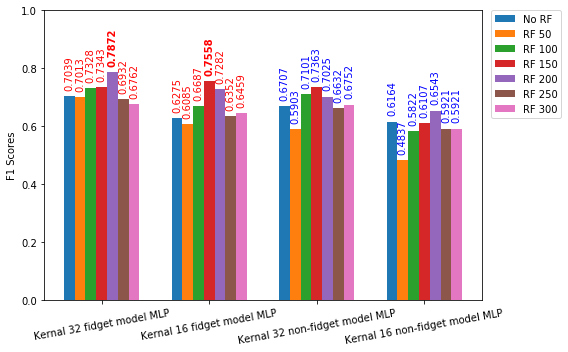}
    \caption{Effects of hyper-parameters. Red denotes models incorporating fidget features and blue for non-fidget models. 
    In general, models with fidget features perform better.
    (Error bars are not shown for better visualization; best performance of each model is in \textbf{bold}). RF+number denotes the number of features selected by Random Forest.}
    \label{fig:effects_of_hyperparameters}
\end{figure}
As shown in Fig.~\ref{fig:label_smoothing}, when other hyperparamters are fixed, label smoothing makes great effects on classification performance. 
Fig.~\ref{fig:label_smoothing} presents the great effect of label smoothing on classification performance when other hyperparameters are fixed.
Though some turbulences exist, the performance increases with higher label smoothing but starts to decrease when smoothing is too much. This is intuitively reasonable because when smoothing is above 0.5, there is less allowed space for model to learn features well.
The results in Fig.~\ref{fig:label_smoothing} shows that label smoothing parameter at 0.4 generally provides good performance, and thus we fixed this value in follow experiments.

We test different numbers of features selected by RF (RF\_num), and different GMM kernel sizes.
Fig.~\ref{fig:effects_of_hyperparameters} shows that the performance is generally worse when RF\_num is low ($<100$) as it results in insufficient information.
However, when RF\_num is high ($\geq250$), redundant features bias the classifier, decreasing performance.

Using 32 GMM kernels achieves better performance than 16 kernels. 
We believe this is due to the way GMM clusters similar per-frame features.
More kernels mean more clusters and thus more predictive information.
However, when kernel size is above 32, the fitting score is large (in GMM lower is better) and therefore increasing beyond 32 will not further improve performance.

\subsubsection{Effects of feature groups}
\begin{figure}
    \centering
    \includegraphics[width=9.2cm]{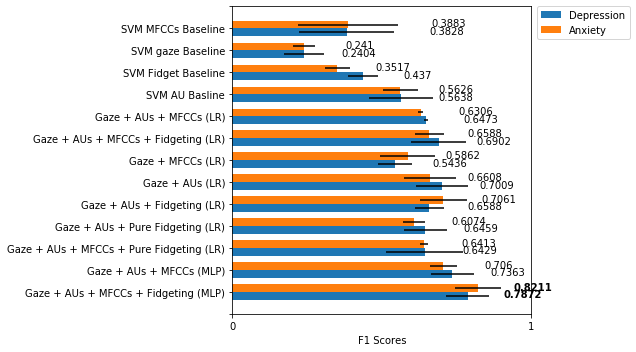}
    \caption{Effects of feature groups and ablation analysis (error bars extend by the standard deviation in either side; best performance is in \textbf{bold}).}
    \label{fig:final_results}
\end{figure}
From Fig.~\ref{fig:final_results}, it is clear that fidget features improve most configurations' performance, but performance decreases slightly without the participant speaking event (presented as ``Pure Fidgeting'' in figure).
This leads us to conclude that the co-occurrence of speaking and fidgeting is relevant for distress detection.

\subsubsection{Ablation Analysis}
Fig.~\ref{fig:final_results} also demonstrates our ablation analysis to help us understand better the important factors in distress classification.
We remove one or two feature groups from our framework and conduct the same experiments.

Without MFCCs features, the performance generally doesn't drop too much in depression and even increases in anxiety. 
This may suggest that MFCCs are not very important in depression and even distractive in anxiety detection. 

AUs have long been proved to be predictive of distress, and, as expected, we see a significant performance reduction when omitting them.

It is interesting to note that fidgeting, with the LR configuration, does not consistently improve performance, but in anxiety, it always boosts the classification results. This allows us to conclude that fidgeting is certainly important in anxiety, but is also predictive in depression when combined with other feature configuration.

\subsection{Fidget detector cross-dataset validation}
To further validate our automatic fidgeting detection approach, we evaluate it on a publicly available dataset from Mahmoud \textit{et al.}\ \cite{mahmoud2013automatic} that has videos of fidgeting behavior along with manual fidgeting labels.

In this dataset, actors perform specific fidgets.
While these fidgets are overemphasized compared to natural fidgets, their core movement is similar.

Segments of the video containing fidgeting are manually labeled in an action-exclusive manner.
That is, the co-occurrence of fidgeting is not labeled.
Given this, we measure the accuracy of our approach in two phases: first, we check that fidgeting, regardless of location, is detected during the periods of manually labeled fidgeting; and second, we calculate the recall for location-specific fidgeting.
Precision would not make sense for location-specific fidgeting, because the detected location may also be fidgeting, while the ground truth only considers one location.

Detected fidgeting segments shorter than 100 frames are excluded to reduce noise.
\begin{table}[!htp]
\centering
    \begin{tabular}{ccccc}
        \multicolumn{4}{l}{\textbf{Step 1}: Detect fidget only}\\
        \toprule
        fidget & precision & recall & f1-score & support\\
        \midrule
        0 & 0.51 &  0.49 &  0.50 &  29440\\
        1 & 0.79 &  0.80 &  0.80 &  69517\\
        \bottomrule
    \end{tabular}
    \\
    \vspace{0.3cm}
    \begin{tabular}{lll}
    \multicolumn{3}{l}{\textbf{Step 2}: Detect specific fidgeting}\\
    (evaluated with recall)\\
    \toprule
    Fidget type& Recall & Support \\
    \midrule
    leg       & 0.784 & 32430  \\
    hand to face       & 0.865 & 10594   \\
    hand to arm        & 0.787 & 12794 \\
    hand cross  & 0.768 & 13699  \\
    \bottomrule
    \end{tabular}
    \vspace{0.2cm}
    \caption{Results of fidget detection on Mahmoud \textit{et al.}'s dataset \cite{mahmoud2013automatic}.}
    \label{tab:result_actor_dataset}
\end{table}
As shown in Table~\ref{tab:result_actor_dataset}, the recall of the non-fidget label is around 50\%, but this due to the fact that the labels are generally assigned to a long continuous segment and do not accurately reflect the actions occurring per-frame. However, the recall of the fidget label is good, achieving 80\%.

Our fidgeting detection approach outperforms the state-of-the-art presented by Mahmoud \textit{et al.} \cite{mahmoud2013automatic} for each fidget type, achieving a recall above 75\% for all fidgeting types.

\section{Conclusion}\label{sec:conclusion}

We introduced a novel audio-visual distress dataset comprising recorded interviews and distress labels based on psychological questionnaires.

We then presented an automated self-adaptor and fidgeting detection system to extract different fidgeting behaviors from real interview videos. We validated our automated approach by evaluating it on a manually-labeled publicly-available fidgeting dataset as well as our newly collected dataset of natural expressions.

Statistical analysis with generic gesture features was carried out, providing interesting insights into the effect of different generic body movements and their correlation with depression labels. 

We also presented a deep learning method that doesn't require a feature search and utilizes the co-occurrence of different multi-modal features. 
We combined our detected fidgeting features with three other modalities, AUs, gaze, and MFCCs, in a multi-modal distress classification pipeline.
This pipeline utilized a Multi-modal Deep Denoising Auto-Encoder to compactly represent the modalities per-frame, a GMM to FV step to represent the features across a whole video compactly, and a random forest to select important features.
Finally, we tested the binary classification of depression/anxiety labels using LR and MLP classifiers.
An ablation study has been carried out to demonstrate the effect of detected fidgeting behaviors 
in predicting signs of psychological distress.


 \section{Limitations and Future Work}
 
Given the limitations of the small dataset we used, more work is required to utilize the fidgeting features as a complementary modality for classification and prediction of psychological distress. Though recruiting participants and interviewing are time-consuming and costly, we are planning to extend our dataset with more videos. In our multi-modal classification experiments, we treated all fidgeting features as a whole. When more data is available, it will be interesting to evaluate the importance of each fidget behavior (e.g., \ hand to arm fidget and hand to hand fidget). In our work, we only focused on depression and anxiety disorders. However, our automatic approach to detecting self-adaptors and fidgeting opens the door for more work to explore the presence of these non-verbal behaviors and measure them quantitatively in other psychological disorders.
\ifCLASSOPTIONcaptionsoff
  \newpage
\fi



%


\bibliographystyle{IEEEtran}
\bibliography{ieee}

%

\begin{IEEEbiography}[{\includegraphics[width=1in,height=1.25in,clip,keepaspectratio]{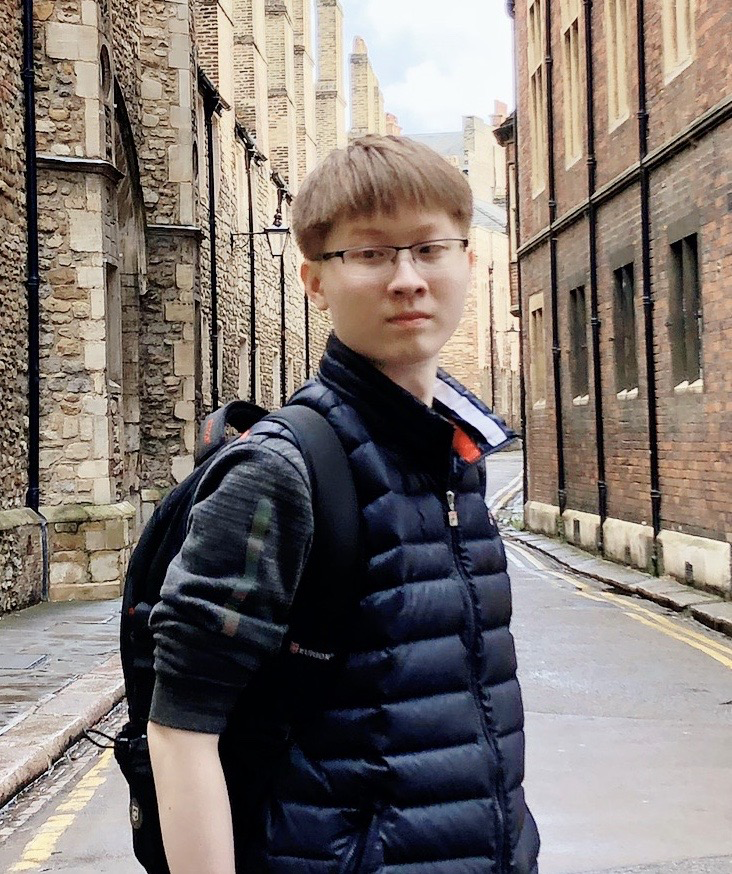}}]{Weizhe Lin}
is a 3rd-year undergraduate student in Information Engineering at University of Cambridge.
He studied Computer Science at Hong Kong University for one year, and then transferred to Cambridge for a 4-year BA/MEng course in Engineering. 
He was awarded a silver medal in ``Future Scientist Award Program'' by Chinese Academy of Sciences and Academy of Engineering.
His research interest spans Natural Language Processing, Hyperspectral Image Processing, and Affective Computing.
\end{IEEEbiography}


\begin{IEEEbiography}[{\includegraphics[width=1in,height=1.25in,clip,keepaspectratio]{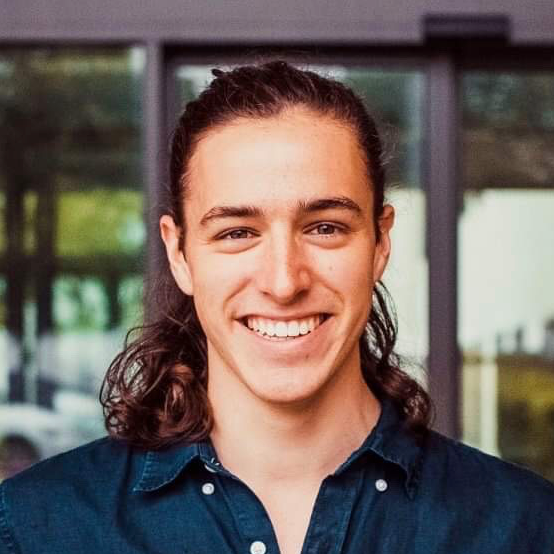}}]{Indigo Orton}
is a PhD student in Computer Science at the University of Cambridge.
From Australia, Indigo completed his Bachelor in Information Technology at Deakin University in Melbourne, Australia.
He then spent a number of years in industry building successful startups in the USA and Australia.
Returning to academia, Indigo received his MPhil in Computer Science from Cambridge in 2019 and later that year commenced his PhD.
\end{IEEEbiography}

\begin{IEEEbiography}[{\includegraphics[width=1in,height=1.25in,clip,keepaspectratio]{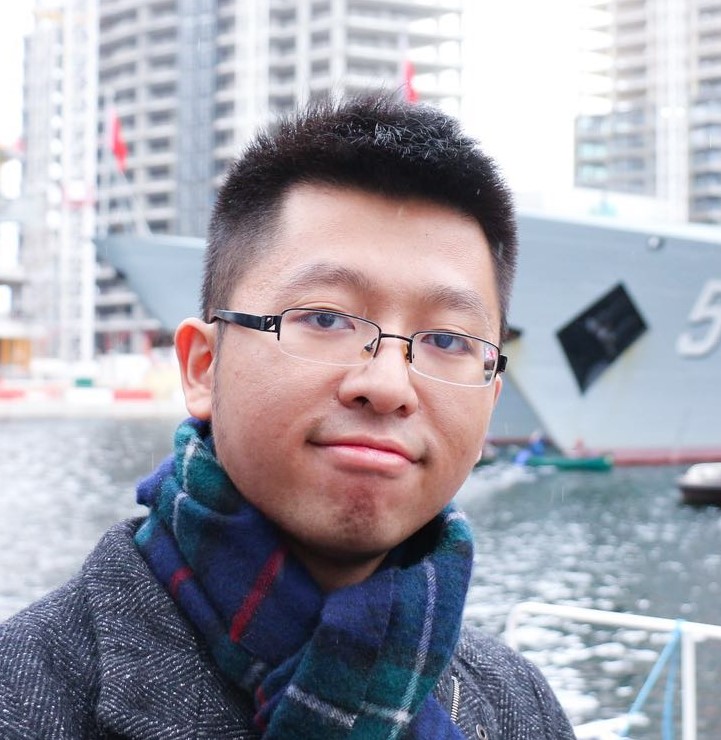}}]{Qinbiao Li}
is a Ph.D. student in the Prorok Lab at the University of Cambridge. Prior to joining Cambridge, he completed an MRes degree in Medical Robotics at Imperial College London, and MEng degree in Mechanical Engineering at University of Edinburgh. His research focus now is mobile robots using Graph Neural Networks (GNN). He has published papers in medical imaging area (IJCARS) and robotics (IEEE Humanoids and AAMAS).
\end{IEEEbiography}

\begin{IEEEbiography}[{\includegraphics[width=1in,height=1.25in,clip,keepaspectratio]{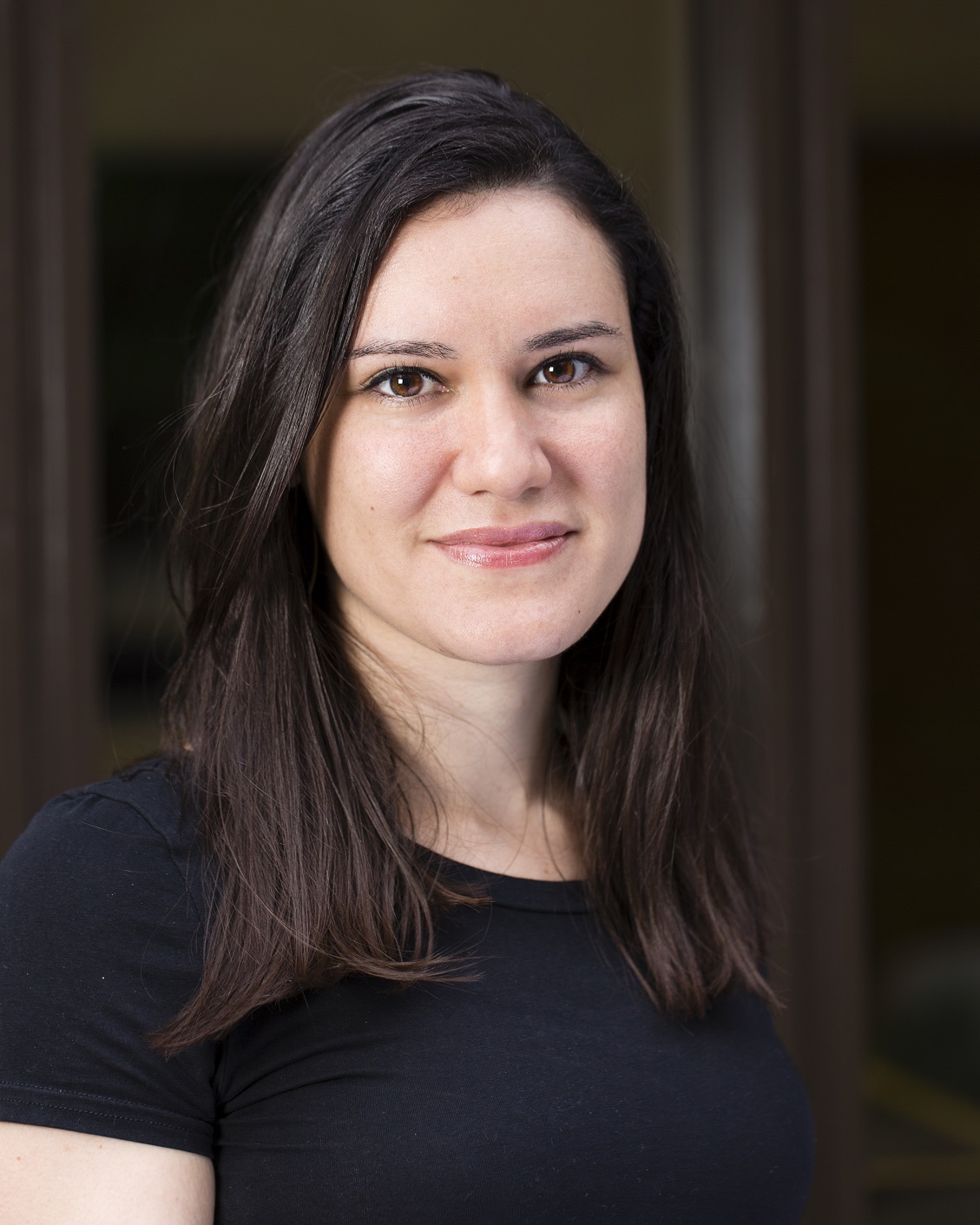}}]{Dr Gabriela Pavarini}
is a Postdoctoral Research Fellow at the Department of Psychiatry at the University of Oxford. Her current research focuses on adolescent mental health, peer-led interventions and ethics of new technologies in psychiatry. She co-designs digital tools such as games and chatbots to facilitate youth engagement in the design and implementation of mental health care.
\end{IEEEbiography}

\begin{IEEEbiography}[{\includegraphics[width=1in,height=1.25in,clip,keepaspectratio]{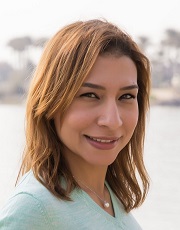}}]{Dr Marwa Mahmoud}
is a Research Fellow of King's College and an Affiliated Lecturer at the Department of Computer Science and Technology, University of Cambridge. Her research focuses on computer vision and machine learning within the context of affective computing, behaviour analytics and human behaviour understanding. 
She is particularly interested in building inference models that tackle challenging real-world problems, usually characterised by data scarcity and noisy signals from multiple modalities. She applies her research in the areas of automotive applications, healthcare, and animal welfare.
\end{IEEEbiography}




\clearpage
\end{document}